\documentclass{article}

    \PassOptionsToPackage{numbers}{natbib}

 \usepackage[sglblindworkshop, final]{neurips_2025}

\workshoptitle{Machine Learning for Systems}

\usepackage[utf8]{inputenc} 
\usepackage[T1]{fontenc}    
\usepackage{hyperref}       
\hypersetup{
    colorlinks,
    citecolor=blue,
    filecolor=black,
    linkcolor=black,
    urlcolor=magenta
}
\usepackage{url}            
\usepackage{booktabs}       
\usepackage{amssymb}
\usepackage{amsmath}
\usepackage{amsfonts}
\usepackage{amsthm}
\usepackage{mathtools}
\usepackage{dsfont}         
\usepackage{nicefrac}       
\usepackage{microtype}      
\usepackage{xcolor}         
\usepackage{graphicx}
\usepackage[inkscapearea=page]{svg}
\usepackage{adjustbox}
\usepackage{subcaption}
\usepackage{multirow}

\makeatletter
\renewcommand{\paragraph}{%
  \@startsection{paragraph}{4}%
  {\z@}{.5ex \@plus .5ex \@minus .2ex}{-1em}%
  {\normalfont\normalsize\bfseries}%
}
\makeatother

\title{Advancing Routing-Awareness in Analog ICs Floorplanning}
\workshoptitle{Machine Learning for Systems}

\author{%
Davide Basso$^{1,2}$ \quad Luca Bortolussi$^{1}$ \quad Mirjana Videnovic-Misic$^2$ \quad Husni Habal$^3$\\
$^1$University of Trieste\quad$^2$Infineon Technologies AT \quad$^3$Infineon Technologies AG\\
\texttt{davide.basso@phd.units.it,lbortolussi@units.it}\\
\texttt{\{mirjana.videnovic-misic, husni.habal\}@infineon.com}
}

\begin{document}

\maketitle

\vspace{-.5cm}
\begin{abstract}
The adoption of machine learning-based techniques for analog integrated circuit layout, unlike its digital counterpart, has been limited by the stringent requirements imposed by electric and problem-specific constraints, along with the interdependence of floorplanning and routing steps.
In this work, we address a prevalent concern among layout engineers regarding the need for readily available routing-aware floorplanning solutions. To this extent, we develop an automatic floorplanning engine based on reinforcement learning and relational graph convolutional neural network specifically tailored to condition the floorplan generation towards more routable outcomes.
A combination of increased grid resolution and precise pin information integration, along with a dynamic routing resource estimation technique, allows balancing routing and area efficiency, eventually meeting industrial standards.
When analyzing the place and route effectiveness in a simulated environment, the proposed approach achieves a $13.8\%$ reduction in dead space, a $40.6\%$ reduction in wirelength and a $73.4\%$ increase in routing success when compared to past learning-based state-of-the-art techniques~\cite{basso_effective_2025}.
\end{abstract}

\section{Introduction}
\label{sec:intro}
Designing the layout of analog integrated circuits (ICs) is a crucial and complex task that requires significant expertise in order to be successfully completed. The susceptibility to noise and parasitics, combined with the need of strict topological requirements, pose significant challenges for layout engineers, often necessitating multiple iterations to achieve an optimal layout that meets performance and design constraints.
The entire process can be subdivided in \emph{floorplanning} and \emph{routing} steps, i.e. placing and connecting the analog devices, respectively. These two phases are inherently interdependent, with the optimization objectives of one stage directly affecting the feasibility and quality of the other. For instance, a compact but unroutable floorplan makes the automation efforts ineffective, highlighting the importance of considering routing metrics early in the layout stage.

Fully automated layout engines, such as MAGICAL~\cite{chen_magical_2021} and ALIGN~\cite{dhar_align_2021}, have proved to be effective in delivering analog circuit layouts by employing analytical or metaheuristic-based floorplanning engines, that connect all the components together typically through the A*~\cite{foead_astar_2021} algorithm. Some efforts have been made to integrate learning-based solutions into these pipelines as well, targeting both the floorplanning~\cite{li_customized_2020, liu_towards_2020} and routing phases~\cite{zhu_geniusroute_2019, chen_reinforcement_2023}. However, these promising approaches have shown limited industrial adoption, highlighting the need for further exploration in this direction.

On the other hand, the combination of relational graph convolutional neural networks (R-GCNs)~\cite{schlichtkrull_modeling_2018} and reinforcement learning (RL)~\cite{sutton_reinforcement_2020} proposed in~\cite{basso_effective_2025} ensures high quality outcomes that can meet industrial expectations when inserted into a real-world working pipeline. For this reason, to increase the routing-awareness of such methodology, we opt to extend its capability with the following enhancements. First, we develop a higher resolution grid to enable more compact and precise device placement, ensuring finer granularity in the layout space. Then, we include detailed information from pins into both the graph representation and RL agent, accounting for more accurate half-perimeter wirelength (HPWL), computed as the half-perimeter of all nets' bounding boxes, and significantly improving the agent’s understanding of the layout problem as a whole. Moreover, a dynamic routing resource estimation based on such new pin information is used to reserve enough space between floorplan elements, nicely accommodating the routing demands in a highly flexible manner. Furthermore, we introduce a revised reward scheme that actively prefers the generation of more routing-friendly floorplans by prioritizing HPWL optimization without sacrificing area efficiency, ensuring improved final outcomes.
Finally, we devise a custom A* analog routing engine, inspired by~\cite{chen_toward_2020, zhang_sageroute_2023}, to generate complete layouts within a simulated environment. The proposed framework outperforms past approaches both in terms of raw placement and routing-related metrics, delivering results which are consistently more routable.

\section{Routing-Aware Floorplanning}
\label{sec:methods}
\begin{figure}
    \centering
    \includegraphics[width=.8\linewidth]{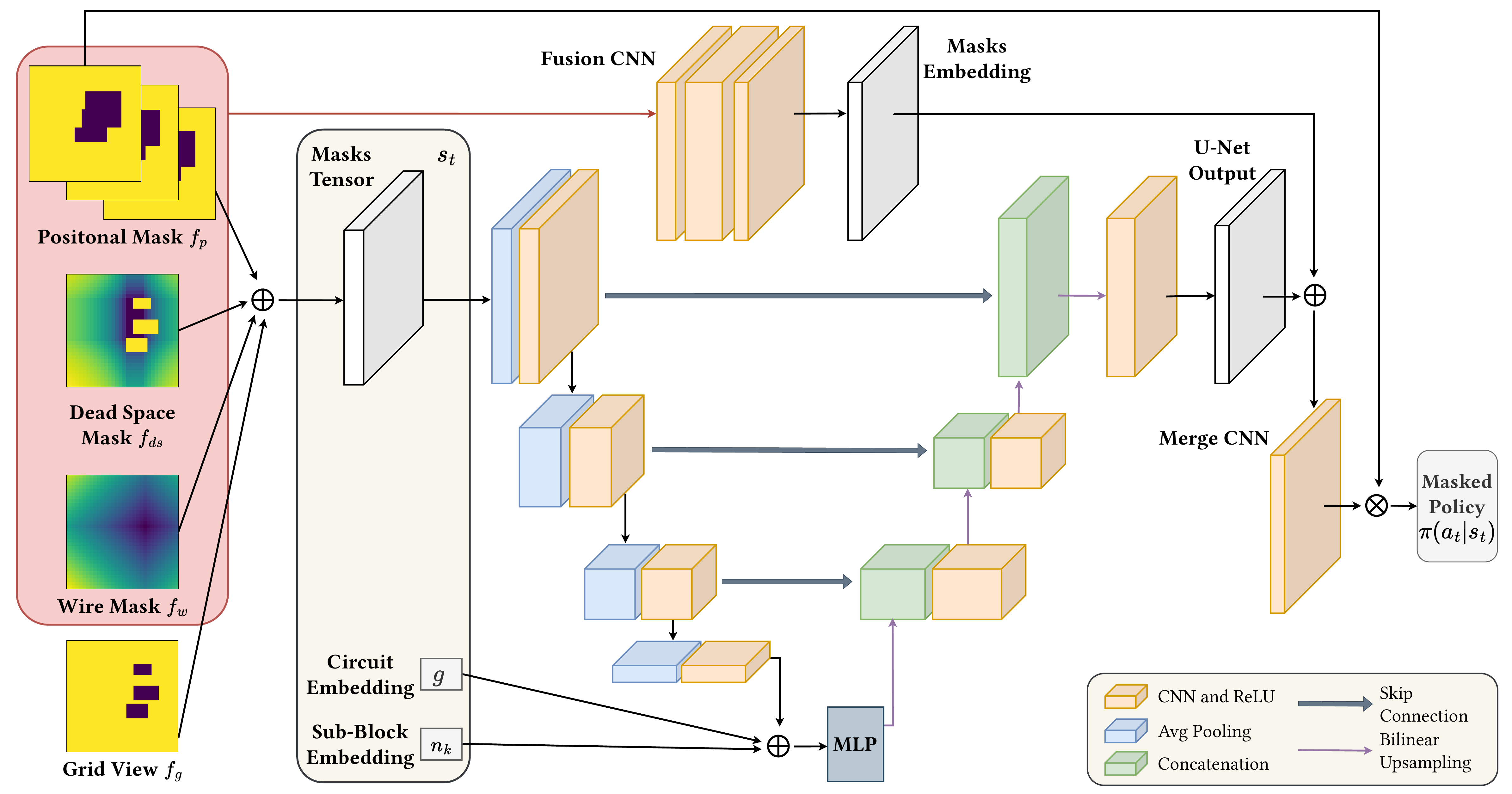}
    \caption{Overview of the RL framework, with a specific focus on the U-Net based policy network.}
    \label{fig:new_policy}
\end{figure}

Floorplanning involves placing transistor, resistor, and capacitor shapes on the IC layout canvas, with the goal of minimizing area and ensuring efficient routing.  Modeled as a Markov Decision Process (MDP), RL has achieved state-of-the-art results in digital layout applications~\cite{lai_maskplace_2022, lai_chipformer_2023, yang_miracle_2024}, leveraging increased capability in exploring large solution spaces compared to metaheuristic techniques.

An overview of the proposed \emph{analog} routing-aware RL floorplanning engine is depicted in Figure~\ref{fig:new_policy}. The state $s_t$ captures both global and local problem information, comprising the R-GCN-generated circuit graph $g$ and current instance $n_k$ embeddings, along with the reward and current grid status masks. Action $a_t$ consists in selecting one of three possible shapes for the current block $b_t$ and placing its lower-left corner in a specific grid cell. The policy network $\pi(a_t|s_t)$ ensures valid actions by applying a mask to prevent device overlaps. A partial reward $r_t$ is defined as the negative change in HPWL and dead space after action $a_t$, while the end of episode reward is defined as the negative weighted sum of floorplan area, HPWL, and the discrepancy from the target aspect ratio.

\paragraph{Pin-Enhanced Circuit Graph}
Circuit netlists can be naturally represented as a graph $G(V,E)$~\cite{lopera_survey_2021}, and for this reason we decide to employ R-GCNs to retain rich circuit and sub-block embeddings, to be then fed to the RL agent for enhancing its problem understanding and generalization capabilities~\cite{basso_effective_2025, amini_generalizable_2022, mirhoseini_graph_2021, le_toward_2023}.
R-GCNs can, by construction, differentiate between diverse type of edges $e_i \in E$ and nodes $v_i \in V$. Electrical connections through a net or topological constraints that hold between components are embedded as edges, while nodes include functional sub-blocks, devices, pins and nets.
An example of the proposed circuit graph representation is shown in Figure~\ref{fig:new_circuit_graph}.
The node feature vectors, edge relationship, and the R-GCN training procedure details are presented in Appendix~\ref{appendix:rgcn}.\looseness=-1

\paragraph{Policy Architecture Improvement}
The granularity of the floorplanning grid in~\cite{basso_effective_2025} is increased from $32 \times 32$ to a $256 \times 256$, enabling higher precision for the analog-specific alignment and symmetry constraints.
To sustain the larger action and state space, we transition from a simple convolutional neural network (CNN)~\cite{lecun_handwritten_1989} to a more sophisticated U-Net~\cite{ronneberger_unet_2015} for the RL policy architecture, similar to the approach in \emph{EfficientPlace}~\cite{geng_reinforcement_2024}.
This update allows for efficient information extraction from visual grid masks, which is combined with R-GCN embeddings to reconstruct action probabilities, as in Figure~\ref{fig:new_policy}, all while maintaining the scalability of the RL agent.
Appendix~\ref{appendix:u_net_design} details the U-Net design.\looseness=-1

\paragraph{Dynamic Routing Resource Allocation}
To avoid routing congestion, we devise a dynamic routing resource (DRR) allocation mechanism, first proposed by~\cite{ou_simultaneous_2013}. The routing space $\lambda_{i,[\mathrm{h,v}]}$ is symmetrically padded around each block based on the number and direction of pins, as well as routing grid constraints. Additional details are provided in Appendix~\ref{appendix:drr}. 

\paragraph{Routing-Driven Reward Design}
The agent's reward at the final time step $T$ is defined as:
\begin{equation}
    \label{eq:reward_func}
    r_T = -\left( \alpha \frac{F_\text{area}}{\sum_{i=1}^m A_i} + \beta \frac{\text{HPWL}}{\text{HPWL\textsubscript{min}}}  + \gamma(\text{Ar}^* - \text{Ar})^2 \right),
\end{equation}
where $F_\text{area}$ is the final floorplan area, $A_i$ denotes the area of the $i^\text{th}$ device, $\text{HPWL\textsubscript{min}}$ is the minimum HPWL value obtained through metaheuristic-based simulations, and $\text{Ar}^*$ and $\text{Ar}$ are the target and current floorplan aspect ratios, respectively.
To prioritize HPWL optimization, a linear lexicographical weighting system is introduced. Instead of solving individual minimization problems sequentially, a computationally efficient alternative assigns weights such that $\alpha_1 \gg \alpha_2 \gg \dots \gg \alpha_m$. Following this idea, the reward weights $\alpha$, $\beta$, and $\gamma$ in Equation~\ref{eq:reward_func} are set to $10$, $100$, and $5$, respectively. A penalty of $-1000$ is further applied to discourage violations of predefined constraints.

\paragraph{Updated Dataset and RL Training Specifics}
The RL agent is trained using a hybrid curriculum learning approach~\cite{zaremba_learning_2014}, starting with smaller circuits and periodically sampling new circuit and constraints. 
The training dataset includes $10$ circuits based on the Infineon $130$nm Commercial Technology node that vary in nature, topology, and functionality. These encompass operational transconductance amplifiers (OTAs), biasing circuits, clock resynchronization circuits, sense amplifiers with RS latches, delay elements, band-gap voltage references with low-power operational amplifiers, and level shifters. Circuit complexity varies, with $3$ to $10$ functional sub-blocks and $9$ to $26$ individual devices.\looseness=-1

\subsection{A* Routing Engine}
\label{subsec:route_place_a_star_routing}
To close the loop between floorplanning and routing and validate the results obtained with the new routing-aware RL agent in terms of routability, we develop an A*-powered rip-up and reroute routing engine, inspired by the promising methodologies employed by both state-of-the-art MAGICAL router~\cite{chen_toward_2020} and SAGERoute~\cite{zhang_sageroute_2023, zhang_sageroute_2024}. 
The prototype, despite being at early stages of its development, allows us to effectively assess the routing awareness of the proposed floorplans, avoiding DRC errors by construction, and supporting multiple metal layers routing. The insights regarding the routing grid, rule checking and routing scheme are presented in Appendix~\ref{appendix:a_star_routing}.

\section{Experimental Results}
\label{sec:results}
We compare the proposed routing-aware floorplanning engine with the state-of-the-art \emph{R-GCN RL} methodology from~\cite{basso_effective_2025}, using Proximal Policy Optimization~\cite{schulman_proximal_2017} to train the RL agent.
We first compare raw placement performance between the proposed approach and both the vanilla and fine-tuned versions of the baseline. Then, we evaluate routing-specific metrics to assess the generation of routing-ready floorplans. The routing-specific results are presented below, while placement-related metrics are detailed in Appendix~\ref{appendix:results}.

\paragraph{Routing Metrics Assessment}
To evaluate routing effectiveness, we applied the A* routing engine to the floorplans generated by both the new routing-aware method and the previous R-GCN RL approach. Key metrics include total routing wirelength, routing failure rate, average iterations for a valid layout, and final via count. The A* algorithm was limited at $35$ rip-up and reroute iterations. As the proposed method does not use routing engine feedback during training, both RL training and $6$ additional circuits, presented in Appendix~\ref{appendix:route_place_floorplanning_results}, were tested to ensure comprehensive evaluation.

\begin{table}
\centering
\caption{Routing-related metrics comparison only successfully routed floorplans. The failure rate is instead computed on all available circuits.}
\label{tab:routing_related_perf}
\resizebox{.8\textwidth}{!}{%
\begin{tabular}{@{}ccccc@{}}
\toprule
Algorithm             &    Subset    & Routing-Aware RL              & R-GCN RL 0-shot      & R-GCN RL 1000-shot          \\ \midrule
Failure Rate (\%)     &   All     & \textbf{9.56}                 & 79.0                 & 83.0                        \\ \midrule
\multirow{2}{*}{Wirelength (\textmu m)} & All & 6496.76 $\pm$ 448.58 & 8732.97 $\pm$ 881.36 & 5637.36 $\pm$ 731.86 \\
                      & Common & \textbf{5742.17 $\pm$ 212.71} & 7107.96 $\pm$ 666.75 & 6264.70 $\pm$ 813.04        \\ \midrule
\multirow{2}{*}{Vias} & All    & 144.16 $\pm$ 7.98             & 140.58 $\pm$ 6.86    & 107.75 $\pm$ 7.53           \\
                      & Common & \textbf{113.48 $\pm$ 5.29}    & 117.51 $\pm$ 4.50    & 113.74 $\pm$ 7.88           \\ \midrule
\multirow{2}{*}{Routing Iterations}                    & All & 3.94 $\pm$ 4.06      & 10.01 $\pm$ 2.46     & 3.64 $\pm$ 2.78      \\
                      & Common & \textbf{3.15 $\pm$ 2.96}      & 10.26 $\pm$ 1.96     & 3.57 $\pm$ 2.83             \\ \bottomrule
\end{tabular}%
}
\end{table}

\begin{table}
\centering
\caption{Performance Comparison On Delay Line Circuit.}
\label{tab:cefu_perf}
\resizebox{\linewidth}{!}{%
\begin{tabular}{@{}ccccccc@{}}
\toprule
Algorithm          & Dead Space (\%)   & HPWL (\textmu m)    & Wirelength (\textmu m) & Vias                & Failure Rate & Runtime (s) \\ \midrule
Routing-Aware RL   & 62.41 $\pm$ 4.82 & 755.54 $\pm$ 60.91  & 14243.0 $\pm$ 2153.25  & 363.04 $\pm$ 33.25  & 0.24         & 48.7        \\
R-GCN RL 0-shot & 78.29 $\pm$ 4.10 & 57035.68 $\pm$ 100092.32 & 15943.62 $\pm$ 15546.75 & 286.12 $\pm$ 298.75 & 1.0 & - \\
R-GCN RL 1000-shot & 77.71 $\pm$ 5.13 & 1026.53 $\pm$ 101.0 & 17715.42 $\pm$ 5243.25 & 297.18 $\pm$ 122.25 & 1.0          & -           \\ \bottomrule
\end{tabular}%
}
\end{table}

\begin{figure}
    \centering
    \includegraphics[width=.8\linewidth]{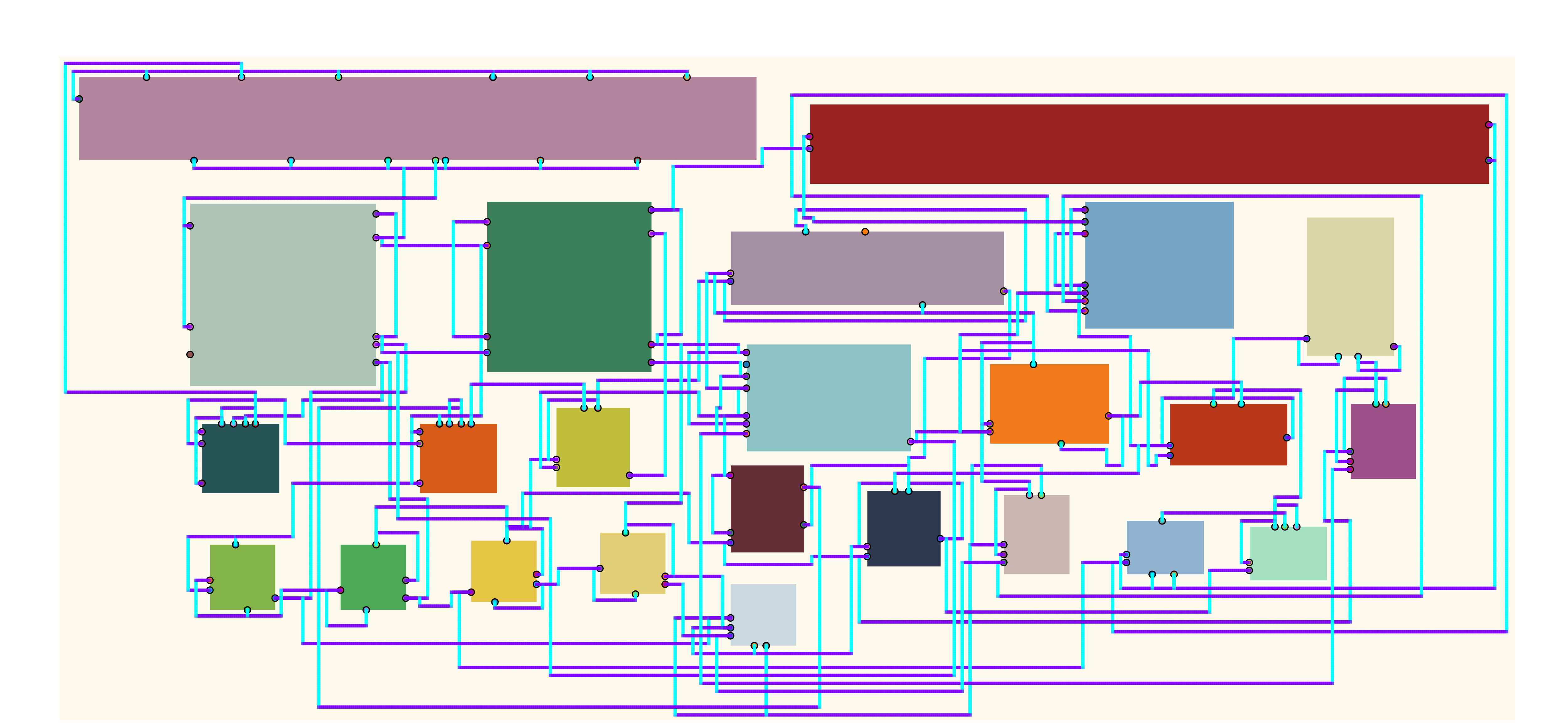}
    \caption{Complete Layout of The Delay Line for Analog Core Signal Timing Circuit.}
    \label{fig:full_layout}
\end{figure}

Table~\ref{tab:routing_related_perf} presents the interquartile mean (IQM) and interquartile range (IQR) for routing metrics, calculated on successfully routed circuits. Given that the routing-aware system successfully routes all available circuits, unlike the other methods, the comparison mainly focuses on metrics derived from circuits routed by all three approaches. The best results are highlighted in bold.
The routing-aware approach outperforms prior methods, achieving a $73.44\%$ reduction in routing failure rate due to precise pin information and HPWL optimization. For circuits routed by all methods, it reduces total wirelength and routing iterations by $8.34\%$ and $11.8\%$, respectively, compared to the R-GCN RL $1000$-shot finetuned method, and by $19.2\%$ and $69.3\%$, respectively, compared to the $0$-shot baseline.

Figure~\ref{fig:full_layout} shows the complete layout of a $24$-block delay line, the largest and most complex circuit in the test set, used in analog core signal timing applications. The routing process benefits significantly from the spacing introduced by the DRR allocation procedure, which, combined with an optimized HPWL during floorplanning, ensures optimal routing paths while adhering to the defined DRC rules.
As reported in Table~\ref{tab:cefu_perf}, previous approaches fail to produce any fully working routing solution. In contrast, the routing-aware solution achieves successful routing in $76\%$ of the cases in just $48.7$s on average.\looseness=-1

\section{Conclusion}
\label{sec:conclusion}
In this paper, we tackled the analog IC layout problem as a whole by enhancing the R-GCN RL floorplanning system with detailed pin information, HPWL-prioritized optimization and a flexible routing resource allocation strategy to reduce congestion and DRC violations.
Moreover, placement precision was improved by increasing the placement grid resolution. Additionally, an A*-based routing engine prototype was developed to ensure DRC-compliant layouts suitable for import into commercial EDA tools.
The evaluations demonstrate the superiority of the routing-aware system, achieving significantly lower routing failure rates and more compact floorplans compared to previous learning-based approaches. Further research will surely involve improvements in the routing engine and how to integrate it in a feedback loop with the RL placement agent.

\clearpage

\bibliographystyle{abbrvnat}
\bibliography{biblio}

\clearpage
\appendix
\section{R-GCN Insights}
\label{appendix:rgcn}
\subsection{Circuit Graph Details}
The feature vector of a device node in the circuit graph, $x_d \in X$, includes the block area, width, height, the number of pins, and a $28$-dimensional one-hot encoding representing the block's functional structure. 
Similarly, the feature vector of a pin node, $x_p \in X$, is characterized by a series of numerical encodings of pin’s direction, the type of net it is connected to (signal, ground, or power), and a one-hot encoding of both the bottom and top metal layers that the pin can accept, which varies according to the technology specifications. 
Finally, the feature vector of a net node, $x_n \in X$, is constructed by concatenating the number of connected devices and pins.

Connections between nodes are established based on the relationships among them. By construction, device nodes and pin nodes are connected through a \emph{device has pin} relationship, while pin nodes are connected to net nodes via the \emph{pin belongs to net} relation. This approach results in a near-tripartite graph structure. We refer to it as ``near-tripartite" because, in addition to the device-pin-net relationships, we need to consider also the constraints between devices that are also represented as edges. 

\subsection{R-GCN Specifics}
The R-GCN architecture consists of two R-GCN layers followed by a node mean aggregation block to generate the graph embedding, which is subsequently passed through two fully connected layers to predict the reward value.

The training dataset includes $91200$ floorplans with corresponding reward values, optimized for area and proxy wirelength metrics using metaheuristic algorithms such as simulated annealing, genetic algorithms, and particle swarm optimization. Circuit sizes range from $3$ up to $24$ blocks and span a variety of designs, including OTAs, bias circuits, drivers, level shifters, clock synchronizers, comparators, and oscillators. To ensure robustness, the dataset is balanced between constrained and unconstrained floorplans.

The supervised learning task minimizes the mean squared error between the predicted and ground truth rewards for input circuit graphs. Training was performed on a single Nvidia A30 GPU, with a total runtime of $46$ minutes. Additional R-GCN architecture and training hyperparameters are detailed in Table~\ref{tab:rgcn_hyperparameters}.

\begin{table}[!h]
\small
\centering
\caption{R-GCN hyperparameters.}
\label{tab:rgcn_hyperparameters}
\begin{tabular}{@{}ll@{}}
\toprule
\textbf{R-GCN Model}       & \textbf{}     \\ \midrule
R-GCN units                & 64            \\
Learning rate              & 4.7e-3        \\
Dropout                    & 0.1           \\
Optimizer                  & ADAM          \\
Weight Decay               & 5.6e-5        \\
$\beta_1, \beta_2$         & 0.9, 0.999    \\ \midrule
\textbf{FC Neural Network} & \textbf{}     \\ \midrule
Input normalization        & tanh          \\
Activation                 & ReLU          \\
Hidden layer neurons       & {[}64,32,1{]} \\
Loss                       & MSE           \\ \bottomrule
\end{tabular}%
\end{table}

\begin{figure}[!ht]
    \begin{subfigure}{.5\textwidth}
      \centering
      \includegraphics[width=.8\textwidth]{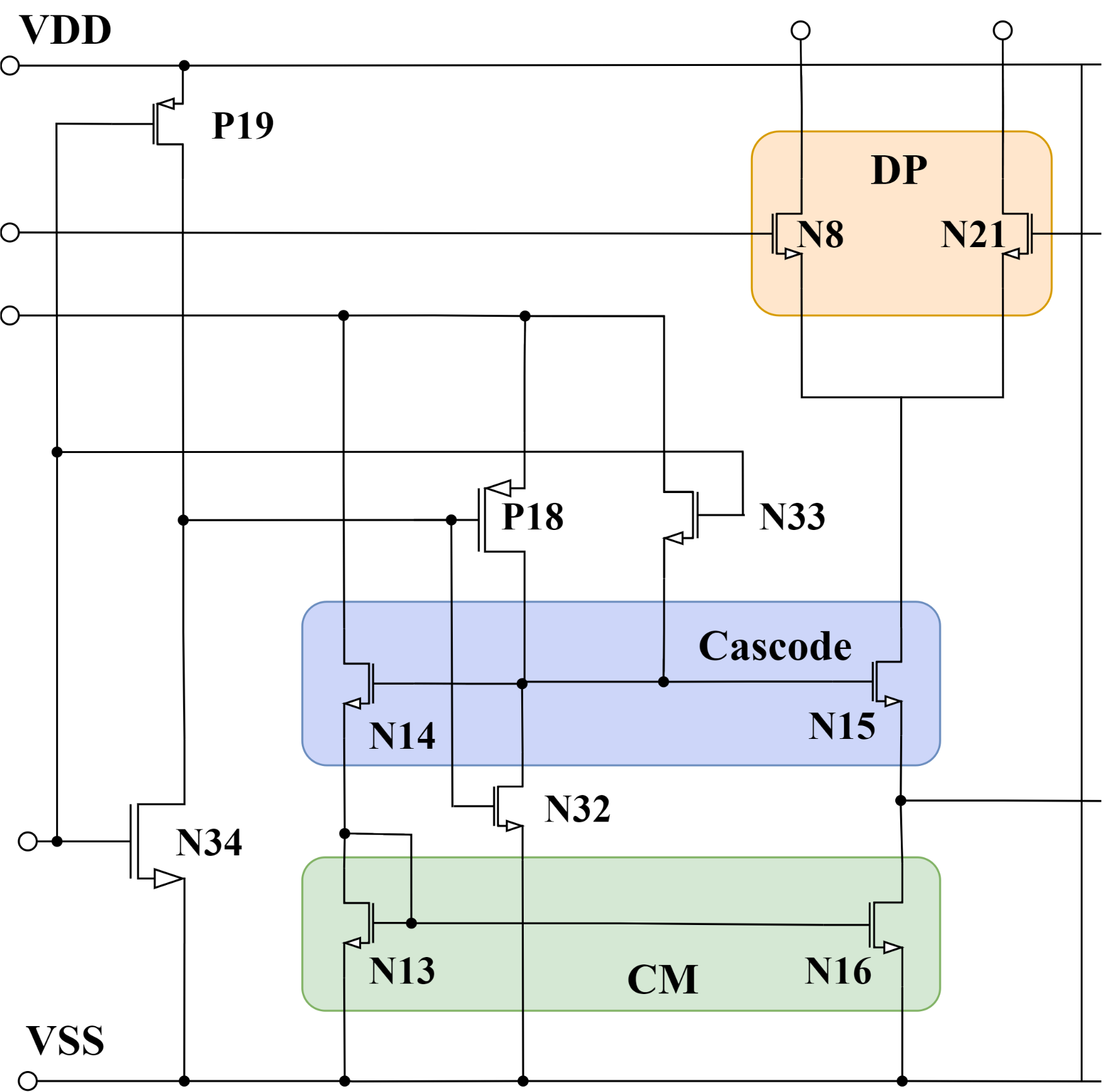}
      \caption{}
      \label{fig:circuit_schematic}
    \end{subfigure}
    \hfill
    \begin{subfigure}{.43\textwidth}
      \centering
      \includegraphics[width=.8\textwidth]{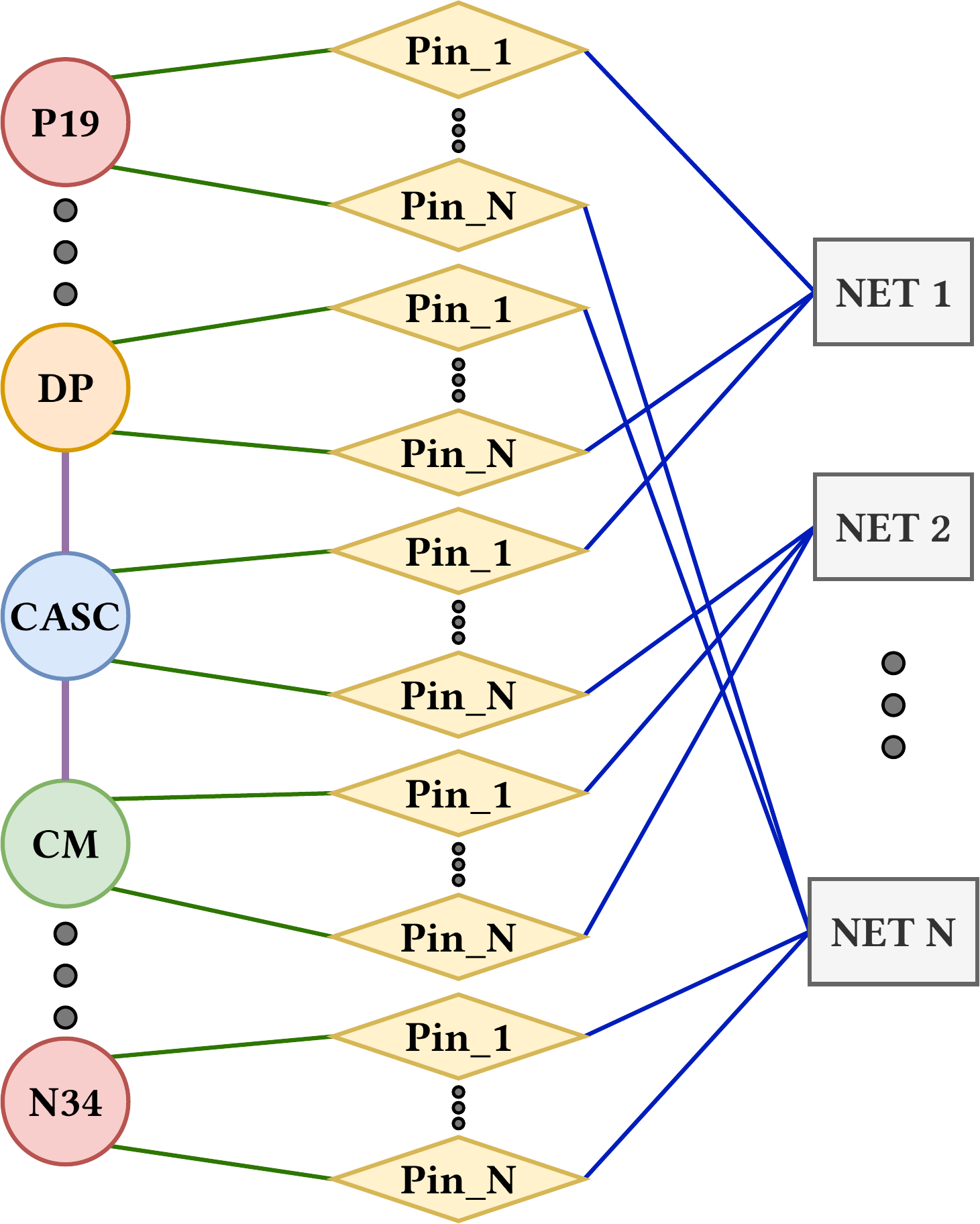}
      \caption{}
      \label{fig:circuit_graph}
    \end{subfigure}
\caption{An 8-block OTA schematic (a) and the near-tripartite circuit graph (b). Violet edges represent the topological constraints applied between recognized structures, green ones denote the \emph{device has pin} relationship, and blue edges connect pins and nets through the \emph{pin belongs to net} relationship.}
\label{fig:new_circuit_graph}
\end{figure}

\section{Enhancing RL Capabilities for Better Routability}
\label{appendix:rl}
\subsection{U-Net Policy Design}
\label{appendix:u_net_design}
Originally designed for medical images segmentation, the U-Net is perfectly suited for efficiently processing visual information coming from the visual masks that are part of the agent's state, $f_g, f_w, f_{ds}, f_p$.
In fact, thanks to its architecture design, which is detailed hereafter, it is possible to process high-dimensional input data without incurring in a significant computational overhead. As a result, despite the increased grid size and the large action space of $|\mathcal{A}| = 256 \times 256 \times 3 = 196608$, employing the U-Net enables effective training and inference.

The U-Net architecture adopts a U-shaped configuration, which can be conceptually divided into three main components: the encoding (or contracting) path, the bottleneck, and the decoding (or expansive) path.

The \emph{contracting path}, located on the left side of the U-Net, serves as a feature extractor for the input image-shaped data while progressively reducing its spatial dimensions. This section typically consists of a sequence of convolutional layers, followed by ReLU activation functions and max-pooling operations.

The \emph{bottleneck}, which forms the central portion of the U-Net, represents the most compact and low-dimensional encoding of the input data. In our setup, this bottleneck incorporates an MLP to efficiently combine not only the information extracted from the contracting path, but also the R-GCN generated circuit $g$, and single block $n_k$ embeddings. Doing so, both global circuit and local block-specific details are integrated into the distilled encoding.

Eventually, the \emph{expansive path}, i.e. the right side of the U-Net, reconstructs the spatial dimensions of the input data through a series of upsampling and convolutional layers. \emph{Skip connections} are used to directly connect layers from the encoding path to their corresponding layers in the decoding section. These help to retain fine-grained spatial details lost during downsampling, enabling the network to reconstruct the spatial structure of the input with high precision.

In order to better guide the RL agent's decision-making process, as the final step of our policy, the state masks are fed to a CNN to eventually retrieve a learned feature vector of such combination. The output of this CNN is then integrated with the probability distribution produced by the U‐Net architecture, using an additional CNN, to produce the final action probabilities for the RL agent.
The value network instead follows the same rationale of the contracting path of the policy network, ultimately outputting the learned value function values through an MLP.

The hyperparameters used for the policy network from Figure~\ref{fig:new_policy} are listed in Table~\ref{tab:new_policy_value_hyperparameters}.

\begin{table}[ht!]
\centering
\caption{RL Policy Model Hyperparameters.}
\label{tab:new_policy_value_hyperparameters}
\resizebox{.9\textwidth}{!}{%
\begin{tabular}{@{}ccccccc@{}}
\toprule
Model Part                     & Layer ID        & Kernel Size  & N. Channels & Stride & Padding & Output Size                    \\ \midrule
\multirow{6}{*}{U-Net Encoder} & E\_CNN\_1       & 3 $\times$ 3 & 8           & 1      & 1       & (8, 128, 128)                  \\
                               & E\_CNN\_2       & 3 $\times$ 3 & 16          & 1      & 1       & (16, 64, 64)                   \\
                               & E\_CNN\_3       & 3 $\times$ 3 & 32          & 1      & 1       & (32, 16, 16)                   \\
                               & E\_CNN\_4       & 3 $\times$ 3 & 32          & 1      & 1       & (32, 4, 4)                     \\
                               & Max\_Pool\_2    & 2 $\times$ 2 & -           & -      & -       & -                              \\
                               & Max\_Pool\_4    & 4 $\times$ 4 & -           & -      & -       & -                              \\ \midrule
\multirow{2}{*}{Bottleneck}    & FC\_1           & -            & -           & -      & -       & (640, 512)                     \\
                               & FC\_2           & -            & -           & -      & -       & (512, 512)                     \\ \midrule
\multirow{6}{*}{U-Net Decoder} & D\_CNN\_1       & 3 $\times$ 3 & 16          & 1      & 1       & (16, 16, 16)                   \\
                               & D\_CNN\_2       & 3 $\times$ 3 & 8           & 1      & 1       & (8, 64, 64)                    \\
                               & D\_CNN\_3       & 1 $\times$ 1 & 3           & 1      & 0       & (3, 256, 256)                  \\
                               & Bilinear\_Up\_2 & -            & -           & -      & -       & (in $\times$ 2, in $\times$ 2) \\
                               & Bilinear\_Up\_4 & -            & -           & -      & -       & (in $\times$ 4, in $\times$ 4) \\ \midrule
\multirow{3}{*}{Fusion CNN}    & F\_CNN\_1       & 1 $\times$ 1 & 8           & 1      & 0       & (8, 256, 256)                  \\
                               & F\_CNN\_2       & 1 $\times$ 1 & 8           & 1      & 0       & (8, 256, 256)                  \\
                               & F\_CNN\_3       & 1 $\times$ 1 & 3           & 1      & 0       & (3, 256, 256)                  \\ \midrule
Merge CNN                      & M\_CNN\_1       & 1 $\times$ 1 & 3           & 1      & 0       & (3, 256, 256)                  \\ \bottomrule
\end{tabular}%
}
\end{table}

\subsection{Dynamic Routing Resources Allocation Details}
\label{appendix:drr}
The routing space, $\lambda_{i,[\mathrm{h,v}]}$ required on either the horizontal or vertical sides of a block $i$, is defined as follows:
\begin{equation}
    \label{eq:DRR}
    \lambda_{i,[\mathrm{h,v}]} = \zeta + |p_{\mathrm{i,[h,v]}}| \cdot \left(\max_{j \in Nets(i)}(\mathcal{W}_{j,\mathrm{spacing}}) + \max_{j \in Nets(i)}(\mathcal{W}_{j,\mathrm{width}})\right),
\end{equation}
where $\zeta$ is equal to the minimum chip canvas placement grid step, $|p_{\mathrm{i,[h,v]}}|$ is the number of vertical or horizontal pins of sub-block $i$, $\max_{j \in Nets(i)}(\mathcal{W}_{j,\mathrm{spacing}})$ is the maximum parallel run spacing among the nets connected to block $i$, $Nets(i)$, and $\max_{j \in Nets(i)}(\mathcal{W}_{j,\mathrm{width}})$ is the maximum wire width for each connection.

Since the output direction of the pins during placement is unknown, the routing space defined in Equation~\ref{eq:DRR} must be padded symmetrically. More precisely, horizontal pins require padding on both the left and right sides, while vertical pins on the top and bottom ones. This approach inevitably results in an overestimation of the actual space needed to connect all components. However, this added room ensures very low routing failure rates, while still achieving well-optimized floorplans in terms of area utilization and HPWL.

\section{Analog-Compliant A* Routing Engine}
\label{appendix:a_star_routing}
\subsection{Routing Grid Graph}
The routing grid is modeled as a $3$‐D graph, $G_r(V_r,E_r)$, representing the entire routing space. $V_r$ denotes the set of possible interconnection points, i.e. vertices, and $E_r$ the set of edges. The edges $(u,v) \in E_r$ can either be horizontal or vertical, depending on the connection type. In fact, horizontal edges correspond to routing track segments within the same metal layer, while vertical ones represent connections between different metal layers, realized physically as vias.

In our setup, the grid step used to interleave two routing vertices is determined by the minimum valid spacing between two physical components specified by the circuit technology.
To efficiently represent $G_r$, since the number of vertices is typically large, an adjacency list is implemented using a hashmap structure, enabling scalable storage and fast lookup of the graph’s components. To further limit the increase in size of $G_r$ we limit the possible metal layers to two options, one for vertical and the other for horizontal tracks.

\subsection{DRC Rules and Constraints}
Analog ICs routing is characterized by a multitude of rules that must be satisfied to meet industrial requirements and practical usability of the final layout. For this reason, we guarantee the compliance to the following fundamental DRC rules:
\begin{itemize}
    \item \emph{Minimum wire length}: Specifies the minimum length for a wire segment in $G_r$.
    \item \emph{Minimum wire area}: Specifies the minimum area for a wire segment in $G_r$.
    \item \emph{End of line spacing}: Defines the minimum required spacing, \emph{eolSpace}, from any other wire or via at the end of a wire segment within the \emph{eolWithin} distance beyond the end of the line. This rule applies only if a parallel wire segment exists within the parallel edge region on the side of the wire segment of interest, defined by the \emph{parWidth} and \emph{parSpace} values.
    \item \emph{Parallel wire spacing}: Specifies the minimum parallel spacing from any other wire or via based on the parallel run length and the maximum width of the two wires. In our scenario, we set it to a minimum fixed valid value. 
\end{itemize}

Obstacle avoidance is also considered in order to ensure that wires do not overlap with pins that are not part of the net of interest or intersect with devices. Moreover, given that analog circuit routing typically involves highly regular shapes to form interconnections between devices, a low‐bending rule is introduced to discourage too tortuous paths, which would eventually lead to an increased number of vias and the possibility to incur in DRC violations.

The wire width is set to the minimum valid value to ensure compliance with DRC rules. However, the system is designed to accommodate varying wire widths for different metal layers, allowing flexibility based on specific routing requirements or technology constraints.

\subsection{The Routing Scheme}
The routing process is based on a negotiated rip‐up and reroute scheme ~\cite{mcmurchie_pathfinder_1995}. Whenever a net violates one of the aforementioned rules or constraints, a history cost associated to all the vertices that have been used to construct the failing wires is accredited before they are removed. Doing so, the A* engine can avoid to visit in successive routing iterations the same nodes that led to some source of errors in the past, incentivizing a different exploration of the grid space.

The net routing order is defined according to an heuristic that allows us to properly rank the nets based on their criticality and size. First of all, nets with more routing failures are prioritized. Then, among nets with the same number of routing failures, nets with higher HPWL are prioritized. If both routing failures and HPWL are identical, the net with more pins is favored. Doing so, not only the final quality of the layout is improved, but also the number of iterations to generate a valid routing.

The output direction of the pins to be connected through a net has to be defined at the very beginning of the routing stage. In fact, pins are typically positioned within a device layout and they need to be projected to one side of the block, either horizontally or vertically, depending on pin's characteristics. 
Since HPWL is an optimal estimator of the true circuit wirelength, the approach we follow is to select the block side for each pin that leads to the smallest increase in the HPWL value of the net to be routed. 
This is achieved by evaluating all the possible combinations of pin-to-block side assignments for the net of interest. While this naive approach increases slightly the computational effort and congestion, it keeps a strong correlation with the smallest estimated wirelength, which in the end is the key goal for an high quality routing solution. In such cases, the earlier padding overestimation around the blocks proves to be highly beneficial.

Once all the endpoints' positions are defined, we decompose each net into a two-pin one by using a MST decomposition.

The A* search follows a bidirectional scheme by alternating a forward and backward search, starting from the source and target pin respectively, in order to improve the speed efficiency of navigating into such a large grid space.
Design rules and constraints are integrated into the node cost function $g(n)$ in a shallow manner. If $n$ is a neighboring node of the current node $k$, the cost function can be expressed as:
\begin{align}
    D(n,k) &= |x_n - x_k| + |y_n - y_k| + c_\mathrm{VIA} \cdot |z_n - z_k| \label{eq:a_star_distance}\\
    g(n) &= g(k) + D(n,k) + c_{\mathrm{DRC}} \cdot \mathds{1}(\mathrm{violation}) + H(n)  \label{eq:rotuing_cost},
\end{align}
where $D(n,k)$ is a Manhattan distance metric that accounts also for a via cost, $c_\mathrm{VIA}$, to discourage excessive layer changes and jagged paths. Instead, $c_{\mathrm{DRC}} \cdot \mathds{1}(\mathrm{violation})$ represents a penalty for DRC violations when visiting the neighboring node, while $H(n)$ is the accumulated history cost of $n$ up to the current reroute iteration. Overall, this formulation ensures that the cost function penalizes undesirable routing behaviors while also promoting efficient and compliant routing paths.
On the other hand, obstacles are directly removed from the routing grid, simplifying the routing process and ensuring that invalid paths are avoided by construction.

The heuristic $h(n)$ used to estimate the cost from the neighboring node to the target goal $t$ is computed as $D(t,n)$. 
In order to gain some speedup in runtime we decided to introduce a \emph{dynamic weighting} mechanism for the cost function as follows:
\begin{equation}
    \label{eq:astar_node_cost_function}
    f(n) = 
    \begin{cases}
        g(n) + h(n) & \text{if }g(n) < h(n), \\
        \dfrac{g(n) + (2\kappa - 1) \cdot h(n)}{\kappa} & \text{otherwise.}
    \end{cases}
\end{equation}
This idea is inspired by the work of Chen et al. in ~\cite{chen_necessary_2021}, which demonstrates how increasing the weight of the heuristic depending on $\kappa$ as the routing path progresses can be used to exploit the proximity to the goal. In our implementation, we set $\kappa = 3$. Doing so, the algorithm prioritizes reaching the target rather than exploring the search space, which on the contrary is preferred during the early stages of the search.

Finally, a \emph{tie-breaking} mechanism is introduced to prevent the A* router from exploring all paths of the same length instead of focusing on a single optimal path. This is achieved by multiplying $h(n)$ by $1+q$, with $q = 0.0001$, which biases the algorithm toward expanding vertices closer to the goal and further reduces unnecessary exploration of the search grid. This approach, combined with the dynamic weighting from before, ensures reasonably fast execution runtimes for the whole pipeline when used both in testing and production environments.

\section{Raw Placement Metrics Assessment}
\label{appendix:results}
The RL agent uses stable-baselines3~\cite{raffin_sb3_2021} and DGL~\cite{wang_dgl_2020} libraries. Due to the increased size of the training dataset, the training process requires slightly more time, taking approximately $3$ days and $20$ hours on a single Nvidia A30 GPU.

The A* routing engine, on the other hand, is implemented in C++, in favor of guaranteeing short runtimes and high computational efficiency. The ankerl~\cite{Leitner-Ankerl_2025} library is used for hashmap operations, the Boost~\cite{schling_boost_2011} library for implementing an efficient R-tree spatial index, and the nlohmann~\cite{Lohmann_2025} library for processing JSON input files.

\subsection{Floorplanning Metrics Assessment}
\label{appendix:route_place_floorplanning_results}
Performances are measured on a test set of $6$ different designs. These include both circuits of similar nature to those in the training set, typically with greater complexity, and entirely different circuit types, such as comparators and oscillators. The number of functional blocks in the test circuits varies from $3$ to $24$, while the total number of individual devices ranges between $5$ and $40$, ensuring enough variety for validating our approach.
No constraints are imposed on any circuit to ensure a fair comparison.

\begin{table}[h!]
\centering
\caption{Comparison of the Routing-Aware Approach Versus R-GCN RL}
\label{tab:aggregated_placement_perf}
\resizebox{.9\textwidth}{!}{%
\begin{tabular}{@{}cccccc@{}}
\toprule
Algorithm &
  \begin{tabular}[c]{@{}c@{}}Finetune\\ Shots\end{tabular} &
  Runtime (s) &
  Dead Space (\%) &
  HPWL (\textmu m) &
  Reward \\ \midrule
Routing-Aware &
  0 &
  0.52 $\pm$ 0.29 &
  {\underline{\textit{57.44 $\pm$ 2.68}}} &
  {\underline{\textit{797.29 $\pm$ 46.51}}} &
  {\underline{\textit{-2.61 $\pm$ 0.13}}} \\
Routing-Aware w/o DRR    & 0    & 0.49 $\pm$ 0.1           & \textbf{37.28 $\pm$ 7.72} & \textbf{673.36 $\pm$ 27.37} & \textbf{-1.65 $\pm$ 0.23} \\
R-GCN RL                 & 0    & \textbf{0.32 $\pm$ 0.12} & 59.78 $\pm$ 11.42         & 1006.75 $\pm$ 167.65        & -7.87 $\pm$ 19.53         \\
R-GCN RL                 & 1000 & 788.82 $\pm$ 1505.63     & \textit{55.98 $\pm$ 9.06} & \textit{749.92 $\pm$ 74.77} & \textit{-1.75 $\pm$ 0.46} \\ \bottomrule
\end{tabular}%
}
\end{table}

Table~\ref{tab:aggregated_placement_perf} showcases the IQM and IQR values obtained on the test set circuits for our new floorplanning methodology runtime, resulting floorplan dead space, HPWL and corresponding reward. We also report the results obtained using the routing-aware RL system without DRR, in order to clearly showcase the enhancement provided by the increased grid size and the capability that our new proposed solution can achieve. The best results are presented in bold, while the second and third best rewards are highlighted in italic underlined and underlined only, respectively.

As expected, the increased grid size delivers the best results across all floorplan quality metrics within less than half a second. More specifically, compared to the updated RL agent without DRR, the R-GCN RL $1000$-shot approach shows increases in dead space and HPWL of $50.2\%$ and $11.4\%$, respectively. For the R-GCN RL $0$-shot case, these increases are even more evident, reaching $60.4\%$ and $49.5\%$ for the same metrics.

In contrast, the routing‐aware system achieves results that nearly match the R-GCN RL $1000$-shot produced floorplans, with dead space and HPWL larger by $2.6\%$ and $6.3\%$, respectively. Nevertheless, the significant runtime improvement, a gain of $150769\%$, makes this new solution our preferred choice.
Moreover, the lower IQR values for every routing-aware solution suggest much more stable performance compared to both previous approaches, making it particularly well‐suited for generating fast and accurate floorplan drafts.


\end{document}